\documentclass[10pt,twocolumn,letterpaper]{article}

\usepackage[final,applications]{wacv}      

\usepackage{graphicx}
\usepackage{amsmath}
\usepackage{amssymb}
\usepackage{booktabs}
\usepackage{multirow}
\usepackage{bbm}
\usepackage{listings}
\usepackage{tablefootnote}
\usepackage{xcolor}
\lstdefinestyle{plaintext}{}
\usepackage{verbatim}

\newcommand{\change}[1]{\textcolor{black}{#1}}
\definecolor{gray}{rgb}{0.5,0.5,0.5}

\definecolor{cvprblue}{rgb}{0.21,0.49,0.74}
\definecolor{Gray}{gray}{0.9}
\definecolor{mauve}{rgb}{0.58,0,0.82}

\usepackage[accsupp]{axessibility}  

\usepackage[pagebackref,breaklinks,colorlinks,citecolor=cvprblue]{hyperref}

\usepackage{orcidlink}

\newcommand{\squeezeup}{\vspace*{-2.5\baselineskip}}
\newcommand{\shrink}{\def\baselinestretch{0.92}\large\normalsize} 
\shrink

\usepackage[pagebackref,breaklinks,colorlinks]{hyperref}

\usepackage[capitalize]{cleveref}
\crefname{section}{Sec.}{Secs.}
\Crefname{section}{Section}{Sections}
\Crefname{table}{Table}{Tables}
\crefname{table}{Tab.}{Tabs.}

\def\wacvPaperID{807} 
\def\confName{WACV}
\def\confYear{2025}

\begin{document}

\title{User-in-the-loop Evaluation of Multimodal LLMs for Activity Assistance}

\author{Mrinal Verghese*\\
Carnegie Mellon University\\
{\tt\small mverghes@andrew.cmu.edu}
\and
Brian Chen*\\
Samsung Research America\\
{\tt\small bc2754@columbia.edu}
\and
Hamid Eghbalzadeh\\
Meta Reality Labs Research\\
{\tt\small heghbalz@meta.com}
\and
Tushar Nagarajan\\
Meta Fundamental AI Research\\
{\tt\small tusharn@meta.com}
\and
Ruta Desai\\
Meta Fundamental AI Research\\
{\tt\small rutadesai@meta.com}
}
\maketitle
\begin{abstract}

Our research investigates the capability of modern multimodal reasoning models, powered by Large Language Models (LLMs), to facilitate vision-powered assistants for multi-step daily activities. Such assistants must be able to~1)~encode relevant visual history from the assistant's sensors, e.g., camera,~2)~forecast future actions for accomplishing the activity, and~3)~replan based on the user in the loop.
To evaluate the first two capabilities, grounding visual history and forecasting in short and long horizons, we conduct benchmarking of two prominent \textbf{classes} of multimodal LLM approaches -- Socratic Models~\cite{socratic} and Vision Conditioned Language Models (VCLMs)~\cite{moon2023anymal} on video-based action anticipation tasks using offline datasets. These offline benchmarks, however, do not allow us to close the loop with the user, which is essential to \change{evaluate the replanning capabilities and }measure successful activity completion in assistive scenarios. To that end, we conduct a first-of-its-kind user study, with 18 participants performing 3 different multi-step cooking activities while wearing an egocentric observation device called Aria~\cite{project_aria} and following assistance from multimodal LLMs. We find that the Socratic approach outperforms VCLMs in both offline and online settings. We further highlight how grounding long visual history, common in activity assistance, remains challenging in current models, especially for VCLMs, and demonstrate that offline metrics do not indicate online performance.






\end{abstract}    
\vspace{-.5cm}
\section{Introduction}

\begin{figure}[t]
    \centering
    \includegraphics[width=0.99\columnwidth]{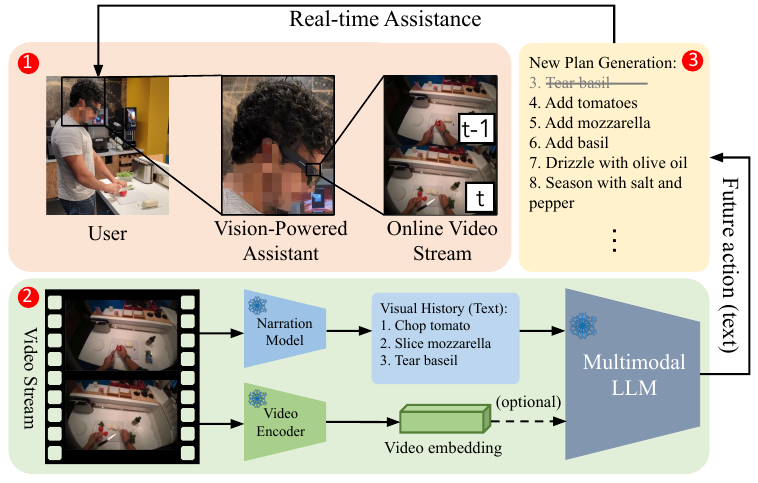}
    \vspace{-0.1cm}
    \caption{\textbf{Online deployment and evaluation of multimodal LLMs for vision-powered assistive systems.}~1)~Users equipped with Aria~\cite{project_aria}, which captures an egocentric video stream of their actions, perform multi-step activities while following assistance from multimodal LLMs.~2)~These models encode the video stream using text-based representations and optionally vision embeddings to predict future actions to complete the activity.~3)~These actions are replanned during execution to provide real-time assistance.}
    \vspace{-.7cm}
    \label{fig:diagram}
\end{figure}

\vspace{-.1cm}
Imagine a vision-powered assistant capable of empowering its users in multi-step daily activities like cooking, assembling, etc. by detecting mistakes and recommending corrections.
Two fundamental capabilities of such an assistant are~a)~the ability to understand task-relevant steps and progress accomplished by the user from the \textit{past} visual observations e.g., video \cite{patel2023pretrained} and~b)~the ability to recommend the next actions the user should take by forecasting and planning of \textit{future} actions \cite{epickitchen21,ego4d}. 
In addition to encoding history and forecasting, such assistants must also account for the user in the loop and re-plan on the fly to ensure successful task execution. With the advent of various vision-language models powered by modern-day LLMs, a natural question is how these models fare towards the aforementioned capabilities. To that end, our work makes two contributions.
First, we perform benchmarking on offline datasets to understand~a)~the effectiveness of different approaches for grounding visual history in multimodal LLMs and~b)~the forecasting capabilities of these approaches in short and long horizons. Second, we test these approaches in online settings to understand whether the performance in offline experiments translates to real-world assistive scenarios with a user in the loop. To the best of our knowledge, our work is the \emph{first} to perform such an online evaluation. 

While a plethora of multimodal LLMs exist today~\cite{huang2023palm,zhao2023antgpt,chen2023videollm,moon2023anymal,li2023videochat,maaz2023video,lin2023videollavalearningunitedvisual,liu2024llava,zhang2023video,cheng2024videollama2advancingspatialtemporal}, they can be broadly differentiated into two categories based on their approach to grounding multimodal inputs 
~--~\emph{Socratic Models} and \emph{Vision-conditioned Language Models (VCLMs)}. The Socratic approach~\cite{socratic} converts visual history into text using pre-trained VLMs such as action or object detectors or narration models. Figure~\ref{fig:diagram} shows the visual history represented as text, e.g., ``chop tomato'', ``slice mozzarella'', etc., obtained from a video narration model for a Caprese salad-making activity. On the other hand, VCLMs such as Flamingo~\cite{flamingo} or LLaVA~\cite{llava} etc. can embed the visual information as continuous embeddings along with text tokens. Specifically, VCLMs split up the available input tokens of their inbuilt LLMs into continuous embeddings from a pretrained vision encoder and text tokens. While VCLMs vary in their training data, underlying LLMs, and vision encoders (Appendix \ref{sup:model_comparison} provides a detailed overview), they all follow a similar general architecture \cite{lin2023videollavalearningunitedvisual,cheng2024videollama2advancingspatialtemporal}. 
Such implicit representation of visual information through continuous embeddings may allow VCLMs to capture details that are hard to encode in text. However, current VCLMs only process a limited number of frames from a long video and tend to uniformly sample these frames~\cite{lin2023videollavalearningunitedvisual}. Consequently, they may miss key-frames and crucial details pertaining to activity assistance. A natural question then is -- \emph{which of these two approaches is more effective for activity assistance?}

To that end, our offline benchmarking evaluates these multimodal LLM approaches using existing video-based action anticipation and visual planning tasks that are representative of challenges in real-world activity assistance~\cite{ego4d,patel2023pretrained}. Specifically, our benchmarks span the spectrum of medium to long visual history and forecasting horizon (Tab.~\ref{tab:offline_benchmarks}). To rigorously compare the Socratic and VCLM approaches, we implement representative models of each using the same LLMs, VLMs, and prompting techniques.
We find that Socratic approaches outperform VCLMs in tasks requiring the grounding of a long visual history, regardless of forecasting horizon. \change{The coarse-level information spanning the entire visual history as captured by Socratic is more important than fine-grained information implicitly captured by VCLMs from a limited set of frames for effective planning in tasks with long visual histories.}
\textcolor{black}{Current VCLMs succeed at providing effective grounding for future action prediction primarily for short to medium visual histories.} Further, such
implicit vision representation in VCLMs is only helpful for planning with small-sized LLMs. 


While these benchmarks evaluate SOTA models in offline settings, it is unclear how such models would perform in online settings where the user is actually performing the activity. Unlike offline datasets, online settings have compute and inference time constraints and require accounting for the user in the loop. The interaction with the user enables the unique opportunity to evaluate the correctness of predicted actions and replanning efficacy toward successful activity completion, which is impossible with offline videos. Specifically, conventional edit distance or success rate metrics for evaluating action plans in action anticipation and planning benchmarks consider matching a single plan present in a given offline video. Instead, online evaluation enables considering all possible action plans that lead to activity success. Albeit, the model must replan at each step, as the user performs the task while grounding the observed task progress from the incoming untrimmed video. 
To test the ability of multimodal LLMs in such settings, we conducted a study with 18 participants performing 3 different multi-step cooking activities while wearing an egocentric observation device called Aria~\cite{project_aria} (Fig.~\ref{fig:diagram}). During activity execution, each participant is instructed to ask for and follow the next action assistance from Socratic and VCLMs at different points in the activity to accomplish it. 
We find that the Socratic approach outperforms VCLMs even in such online settings. Effective online assistance requires identifying task-relevant steps from long, untrimmed, and unsegmented videos while ignoring distractors. \change{Akin to offline settings, the text-based representations of visual history in Socratic models appear better suited to capture such information in comparison to the implicit representations in VCLMs}. Lastly, a head-on comparison between online and offline metrics in our study also highlights that offline metrics such as mean Intersection over Union (mIoU) are an inflated measure of online performance. 


\vspace{-.1cm}
\section{Related Work}
\vspace{-.1cm}

\noindent{\textbf{LLMs for multimodal reasoning.}} Inspired by the capabilities of LLMs, Zeng et al.~\cite{socratic} pioneered the Socratic approach to leverage LLMs for multimodal reasoning. Recently, Palm~\cite{huang2023palm} and AntGPT~\cite{zhao2023antgpt} have employed a similar approach using LLMs to anticipate future actions in videos. 
Specifically, they transform videos into text using narration models like BLIP-2, action recognition models~\cite{zhao2023antgpt}, or a combination of the two~\cite{huang2023palm}. The LLM is then used to model the text sequence to predict future action in a sentence completion fashion \cite{GPT}. However, these approaches have only been evaluated on offline video datasets. In this work, we evaluate such models in an online manner by deploying them in a real-world assistance setting.

\noindent{\textbf{Vision-conditioned Language Model (VCLM).}}
Instead of a text-based representation of vision modality for LLM-based reasoning in vision tasks, another approach is to have a unified model that combines visual and linguistic information by aligning these modalities. Examples of these models include Flamingo \cite{flamingo}, OpenFlamingo \cite{openflamingo}, Palm-E \cite{palme}, BLIP-2 \cite{blip2}, InstructBLIP \cite{instructblip}, LLaVA \cite{llava}, IDEFICS \cite{idefics}, MiniGPT-4 \cite{minigpt4} and many more \cite{otter,mplug,multimodalGPT,lamaadapterv2, su2023pandagpt,lyu2023macaw}. These models are generally fine-tuned using large-scale datasets containing multimodal data \cite{llava, m3it} and are evaluated on image captioning \cite{xu2015show} and visual question answering (VQA) tasks \cite{vqa, visdial, vision-language-navigation}. 

Building on these, VideoLLM\cite{chen2023videollm}, AnyMAL \cite{moon2023anymal}, VideoChat-Embed \cite{li2023videochat}, Video-ChatGPT \cite{maaz2023video}, Video-LLaVA \cite{lin2023videollavalearningunitedvisual}, LLaVA-NeXT \cite{liu2024llava} and Video-LLaMA \cite{zhang2023video,cheng2024videollama2advancingspatialtemporal} fine-tuned VCLMs for a set of video tasks. A nuanced overview of these models can be found in Tab.~\ref{tab:vclm_models} in Appendix \ref{sup:model_comparison}. Keeping the sparsity of available annotated data in assistance scenarios in mind, we specifically focus on few-shot VCLMs without any task-specific finetuning for future action prediction in videos.
Note that various other multimodal multitask transformer models such as GATO~\cite{reed2022generalist} operate on multimodal tokenized input/output from various modalities such as text, image, video, robot actions, etc., for planning. However, we focus on multimodal models that use LLMs as a backbone.

\vspace{-.1cm}
\section{Multimodal LLM Approaches}
\label{sec:offline_mm_llm}
\vspace{-.1cm}
Activity assistance requires grounding information from untrimmed video history for future action prediction. If the visual history is appropriately represented, the future action prediction task can be framed as a sentence completion task using an LLM~\cite{llama,llama2}. We evaluate two predominant approaches to represent visual history for such reasoning with LLMs to predict future actions. Figure \ref{fig:inference}a shows an overview of our Socratic and VCLM models.
 
\noindent\textbf{Socratic model.} The main idea behind the Socratic approach is to \change{use pretrained vision-language models (VLMs) to} convert non-textual modalities into text for a downstream LLM. One could extract and represent different task-relevant information from the untrimmed video history as text, such as objects, actions, and open-set narrations describing the events in the video. We find that such narrations tend to be a superset of various contextual information that could be extracted from video, including objects and actions. Appendix~\ref{sub_additional_ex} provides this quantitative comparison for LTA.  Instead of using objects, actions, and narrations, we choose only narrations from a video narration model to represent visual history as text in our Socratic models, akin to various existing works~\cite{huang2023palm,socratic}. As an example, Socratic models would represent visual information corresponding to visual history in a curry-making activity such as ``Add oil in pan'', ``Add onions in pan'' etc.

\noindent\textbf{Vision-conditioned language models (VCLMs).} These models embed the visual modality as continuous tokens that can be passed as input to an LLM along with text tokens. A linear~\cite{llava, moon2023anymal} or a non-linear projection layer~\cite{patel2023pretrained} is fine-tuned to align these continuous tokens with the embedding space of text tokens for a given LLM. Finally, the LLM backbone is often fine-tuned on a multimodal instruction dataset. \cite{moon2023anymal,cheng2024videollama2advancingspatialtemporal,lin2023videollavalearningunitedvisual,liu2024llava}
Thus, \change{unlike Socratic Models, VCLMs can process both embedded visual information and text. Such implicit representation may allow models to capture fine-grained visual information e.g., ``state of the fried onions'' while making a curry, which might decide if the user should stir more or add the next ingredient.}  

\change{VCLMs typically split up the available tokens in their input context to their inbuilt LLM into continuous embeddings and text tokens, with continuous embeddings coming from visual encoders. The encoders in current SOTA VCLMs use limited and uniformly sampled frames from input videos for video tasks.
Most VCLMs process between 8 to 16 frames \cite{lin2023videollavalearningunitedvisual,liu2024llava,cheng2024videollama2advancingspatialtemporal} 
which may be ineffective in our benchmarks which require grounding on average $>$500 frames corresponding to multiple task-relevant steps. 
To ensure that VCLMs could be applied to our benchmarks, we use both text tokens and continuous embeddings to encode the visual history in our VCLMs.\footnote{This is in contrast to VCLMs used for image captioning and VQA tasks, where visual information is only encoded using the continuous embeddings~\cite{flamingo}.}}
Appendix~\ref{sub_additional_ex} shows an ablation comparing VCLMs that only use continuous embeddings for the encoding history with those that use both continuous embeddings and text tokens in LTA.

\begin{table}[t]
\centering
\resizebox{1\columnwidth}{!}{%
\setlength{\tabcolsep}{2pt} 
\begin{tabular}{@{}cccccc@{}}
\toprule
\textbf{Task}                                                                   & \textbf{Dataset} & \textbf{Metric}                                                                   & \textbf{\begin{tabular}[c]{@{}c@{}}Visual \\ history\end{tabular}}                & \textbf{\begin{tabular}[c]{@{}c@{}}Forecasting \\ horizon\end{tabular}} & \textbf{\begin{tabular}[c]{@{}c@{}}Prediction\\ space\end{tabular}}      \\ \midrule
\begin{tabular}[c]{@{}c@{}}Visual Planning \\ for Assistance (VPA)~\cite{patel2023pretrained}\end{tabular} & CrossTask~\cite{zhukov2019cross}        & \begin{tabular}[c]{@{}c@{}}Mean Accuracy$\uparrow$,\\ Mean IoU$\uparrow$,\\ Success Rate $\uparrow$\end{tabular} & \begin{tabular}[c]{@{}c@{}}Medium\\ (3-4 actions)\end{tabular} & \begin{tabular}[c]{@{}c@{}}Medium\\ (3-4 actions)\end{tabular}          & \begin{tabular}[c]{@{}c@{}}  118 actions\end{tabular}   \\
\begin{tabular}[c]{@{}c@{}}Long-term \\ Action Anticipation (LTA)~\cite{ego4d}\end{tabular}  & Ego4D~\cite{ego4d}            & Edit Distance $\downarrow$                                                                    & \begin{tabular}[c]{@{}c@{}}Long\\ ($>=$ 8 actions)\end{tabular}                   & \begin{tabular}[c]{@{}c@{}}Long\\ (20 actions)\end{tabular}             & 
\begin{tabular}[c]{@{}c@{}}115 verbs, 478 nouns\\ 3542 actions 
\end{tabular} 
\\ \bottomrule
\end{tabular}%
}

\caption{\textbf{Offline benchmarks.} We consider benchmarks spanning the spectrum of medium to long visual history and forecasting horizon, which capture the needs of real-world vision-powered activity assistants. \change{Note that we only consider feasible actions in our prediction space instead of all possible combinations of verbs and nouns.}}
\squeezeup
\vspace{.25cm}
\label{tab:offline_benchmarks}
\end{table}
\vspace{-.1cm}
\section{Offline Benchmarks}
\label{sec:offline_benchmarks}
\vspace{-.1cm}
Our goal is to make progress toward vision-powered assistants that can reason about their user's context from visual input, such as the user's progress in daily activities, and provide relevant recommendations on future actions. Various action anticipation benchmarks previously proposed by the research community also require such reasoning capabilities. Hence, we choose them to evaluate the two prominent categories of SOTA multimodal LLM approaches. 

\subsection{Benchmark Tasks}
\vspace{-.1cm}
While a plethora of video-based action anticipation benchmarks exist~\cite{nagarajan2020ego}, we choose two representative ones such that they cover the space of medium to long visual history and medium to long forecasting horizon -- the settings closest to activity assistance in real-world vision-powered systems\footnote{We exclude anticipation benchmarks like EpickKitchens~\cite{epickitchen21} as they consider short visual histories (1s) and are not aligned with real-world activity assistance.}
Specifically, we choose Long-term action Anticipation (LTA) from Ego4D~\cite{ego4d}, and Visual Planning for Assistance (VPA) task on the CrossTask dataset from~\cite{patel2023pretrained}. We blurred faces from the CrossTask videos prior to use. As summarized in Table~\ref{tab:offline_benchmarks}, LTA focuses on predicting a sequence of future actions 
with a length of $Z=20$ after grounding a long untrimmed visual history corresponding to approximately 8 or more actions. Compared to LTA, VPA on CrossTask operates on a medium-range untrimmed visual history corresponding to 3-4 actions for medium-horizon forecasting of $Z=3\sim4$ future actions.\change{While not part of the original benchmark, we also look at LTA with $Z=5$ to help disambiguate the challenges of long-history and long-horizon in prediction.} The output is mapped to a closed set of verbs, nouns, and actions, i.e., (verb, noun) pairs in each benchmark. We use the same evaluation metrics as were proposed by the original benchmarks for consistency with prior work. The predicted action sequences in LTA are evaluated using the edit distance. VPA is evaluated with action prediction accuracy at each step (mean accuracy), order-agnostic mean Intersection over Union (mIoU), and a strict order-respecting metric Success Rate for the predicted sequence (defined as in~\cite{patel2023pretrained}).


\vspace{-.1cm}
\subsection{Experiment Setup}
\label{sec:experiment_setup}
\vspace{-.1cm}

\begin{figure*}[t]
    \centering
    \includegraphics[width=.85\textwidth]{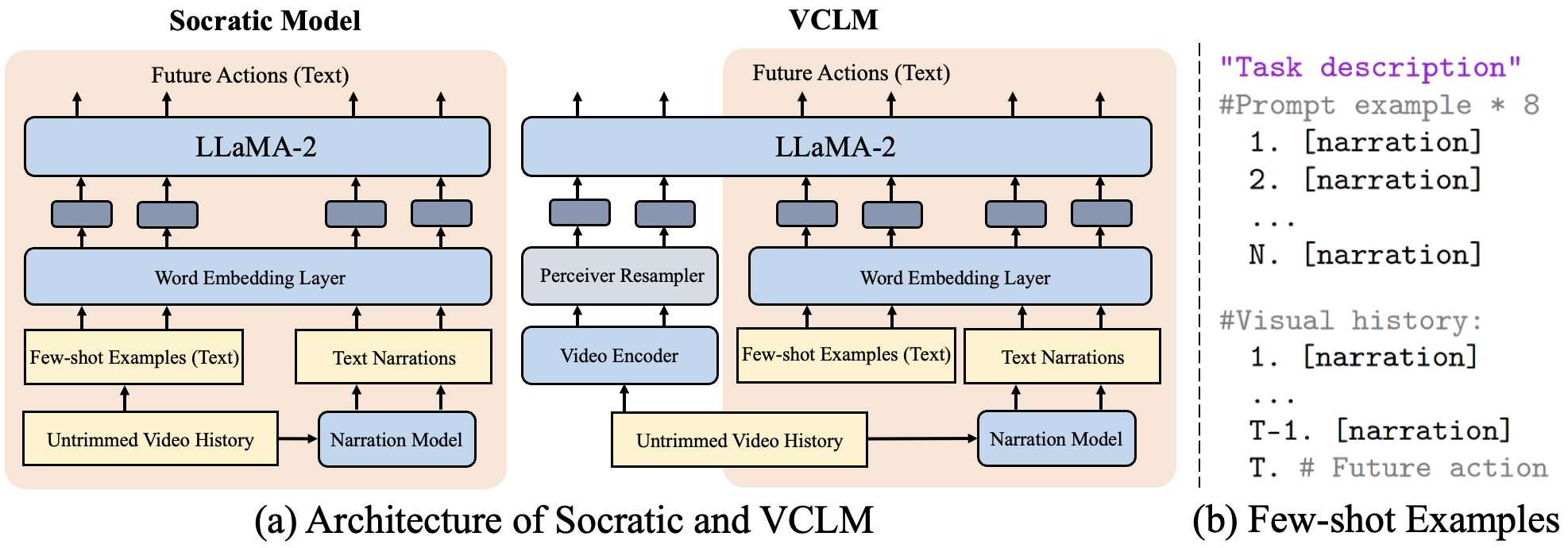}
    \caption{\textbf{Architecture for Socratic and VCLM with few-shot examples.}
     We perform few-shot inference of future actions in our benchmark tasks using untrimmed video history as input. Both Socratic and VCLM convert the video history into text narrations using video narration models (as highlighted in the shaded box). Apart from the narrations, VCLM also embeds the video history as continuous embeddings using a video encoder and Perceiver sampler~\cite{moon2023anymal}. The output is a sequence of natural language sentences, which are then mapped to a closed set of actions for each benchmark.
    }
    \vspace{-.5cm}
    \label{fig:inference}
\end{figure*}

\change{Figure~\ref{fig:inference} shows an overview of Socratic and VCLM models used in our experiments. Both our Socratic and VCLM implementations use the same video narration model to encode the visual history as text. For continuous visual representations, our VCLM model follows the popular projector-based architecture from recent literature~\cite{li2023videochat,lin2023videollavalearningunitedvisual,cheng2024videollama2advancingspatialtemporal,liu2024llava} and consists of a video encoder~\cite{wang2022internvideo} attached to a pretrained LLM with an adapter layer. Both Socratic and VCLM use instruction-tuned 13B and 70B Llama2~\cite{llama2} as their llm backbone, which has a context length of 2048 tokens. While Socratic models use all 2048 of these tokens for text-based representation of history, VCLMs use 256 tokens out of 2048 for continuous visual embeddings of the history.} Since our goal is to understand the capability of these SOTA multimodal LLMs for real-world assistive systems where in-domain data is scarce, we consider these models in the in-context regime only. Appendix \ref{sup:model_comparison} has more information on Socratic and VCLM model variants and our selected implementations within each of these classes of models.

\noindent \textbf{Narrations for visual history.}
To obtain the text-based representation of visual history for Socratic and VCLMs, we leverage video captioning models. Specifically, we used LaViLa \cite{zhao2023lavila}) pre-trained on the Ego4D\cite{ego4d} dataset for Ego4D LTA and video encoder setup following previous work~\cite{patel2023pretrained} for VPA. 
We ensure the best performance in terms of grounding for these multimodal LLMs by~a)~using fine-tuned video narration models and~b)~by using ground-truth segments of the visual history following previous work~\cite{huang2023palm, zhao2023antgpt} in LTA and segments from a finetuned segmentation model following previous work~\cite{patel2023pretrained} in VPA.


%

\noindent \textbf{Prompts for in-context learning.}
For each benchmark task, we generate a set of in-context examples by running the narration model on full, segmented videos from the training set. At test time, we use retrieval-based prompting following \cite{huang2023palm} to select a set of in-context examples that are semantically similar to the narrations from the visual history of the input video. Our final LLM prompt consists of a prompt describing the task of predicting future actions, the set of in-context examples, and the narrations corresponding to the input video's history as shown in Figure \ref{fig:inference}(b). 
For VPA, following previous work~\cite{patel2023pretrained}, we also include the goal of the activity in the video in the prompt by concatenating in front of the narration history. All tasks use 8 in-context examples except where the number of tokens exceeds the LLM's context window. 

\noindent \textbf{Mapping free-form predicted sentences to closed-set action classes.} The output of our multimodal LLMs consists of free-form sentences containing predicted future actions. We map the predicted sentences into the closed-set noun and verb classes for each benchmark by finding the closest word in our class label vocabulary to each word in the output sentence based on the cosine similarity in the SentenceBERT embedding space \cite{thakur-2020-AugSBERT}.

\noindent \textbf{Baselines.} 
We compare VCLMs and Socratic approaches with a set of existing LLM-based and supervised approaches (Tables \ref{tab:long_term_results} and \ref{tab:vpa}). We select these baselines based on the availability of code or results in the Ego4D LTA v1 validation set \cite{ego4d} and the VPA task \cite{patel2023pretrained}. \textbf{LTA baselines.} We include the supervised baseline provided by the Ego4D paper \cite{ego4d} to represent the vision-only approach (not using LM) and a finetuned LLM-based baseline VLaMP \cite{patel2023pretrained} \footnote{We choose VLaMP over VideoLLM~\cite{chen2023videollm} as our finetuned LM baseline because of the availability of code for the former. VideoLLM also only provides results on LTA's test set, preventing a direct comparison on the v1 set.}. Lastly, we consider few-shot LLM-based approaches -- AntGPT \cite{zhao2023antgpt} and Palm \cite{huang2023palm}, which have shown strong performance on the Ego4D LTA task. AntGPT utilizes the activity goal inferred by Llama2-Chat-13B\cite{llama2} and the history of recognized actions in the video history as prompts for Llama2-7B to predict future actions. Likewise, Palm uses the history of recognized actions and narrations from the video history as prompts to GPTNeo-1.3B \cite{gpt-neo} for future action prediction. \change{\textbf{VPA baselines.} Akin to LTA, we provide a supervised non-LM (supervised Ego4D model~\cite{ego4d, patel2023pretrained}) and a fine-tuned LLM-based approach (VLaMP~\cite{patel2023pretrained}) as baselines for VPA. In addition to the Ego4D supervised baseline, which uses a SlowFast video encoder followed by classification heads, we also provide a random action prediction and an LSTM-based supervised baseline called DDN~\cite{chang2020procedure} from the VPA paper~\cite{patel2023pretrained}. }


\subsection{Quantitative Results}
\vspace{-.1cm}



\change{\noindent \textbf{Text-based representation of visual history is more effective than implicit representation when encoding long visual history.}} 
We find that the Socratic approach outperforms VCLMs for predicting actions in Ego4D LTA irrespective of the LLM size (Table \ref{tab:long_term_results}). \change{Our results suggest that the visual embeddings used by VCLMs are less amenable for encoding long visual histories. Specifically, the implicit information contained in these} visual embeddings does not add much beyond the text-based representation of visual history \change{for future prediction tasks that require grounding longer visual histories}.  
\change{This finding is consistent across different future prediction horizons as highlighted by $Z=5$ and $Z=20$ results.}
\change{Recall that irrespective of the history lengths, our Socratic and VCLMs have a fixed context window. While Socratic models use the entire context to encode text, VCLMs use $12.5\%$ of their available tokens in the context window for visual embeddings. For long-history tasks, our results suggest that it is better to use the available context window to encode the history as text rather than devoting tokens to capture implicit information.}


\begin{table}[t]
\centering
\scriptsize
\resizebox{\columnwidth}{!}{
\setlength{\tabcolsep}{2pt} 
\begin{tabular}{lcccllllll}
\toprule
\multirow{2}{*}{\textbf{Model}} &\multirow{2}{*}{\textbf{Visual History}}  & \multirow{2}{*}{\textbf{Visual Encoder}}  &   \multicolumn{3}{c}{ED@(Z=5)$\downarrow$} & \multicolumn{3}{c}{ED@(Z=20)$\downarrow$} \\ 
\cmidrule(l){4-9} 
 &    &  & Verb   & Noun    & Action & Verb   & Noun    & Action\\ \midrule
\multicolumn{1}{l}{AntGPT\cite{zhao2023antgpt}} & T  & CLIP & - & - & - & 0.756  & 0.725   & -       \\ 
\multicolumn{1}{l}{Palm*\tablefootnote{We include Palm$^*$ as the reproduced version of Palm~\cite{huang2023palm}.} \cite{huang2023palm}}&  T  & Blip2 & - & - & - & 0.732  & 0.812   & 0.958    \\ 
\multicolumn{1}{l}{Socratic 13B} & T  & LaViLa & 0.689 & 0.681 & 0.919  & 0.731 & 0.732     &   0.929     \\ 
\multicolumn{1}{l}{Socratic 70B} & T  &LaViLa & \textbf{0.683} & \textbf{0.661} & \textbf{0.917} & \textbf{0.726}   &   \textbf{0.712}  & \textbf{0.928}        \\ 
\midrule
\multicolumn{1}{l}{VCLM 13B} & V+T  & Internvideo+LaViLa & 0.698 & 0.685 & 0.925 &  0.740 &  0.751   &  0.932     \\ 
\multicolumn{1}{l}{VCLM 70B} & V+T  & Internvideo+LaViLa  & 0.696 & 0.669 & 0.923 &  0.739 & 0.731   & 0.931     \\ 
\midrule
\multicolumn{1}{l}{\color{gray}{Ego4D (supervised) \cite{ego4d}}} & \color{gray}{V} & \color{gray}{SlowFast} & \color{gray}{-}  & \color{gray}{-} & \color{gray}{-}   & \color{gray}{0.745}  & \color{gray}{0.779}   & \color{gray}{0.941}    \\ 
\multicolumn{1}{l}{\color{gray}{VLaMP (supervised) \cite{patel2023pretrained}}}& \color{gray}{V+T}  & \color{gray}{S3D}  & \color{gray}{-}  & \color{gray}{-} & \color{gray}{-} & \color{gray}{0.730}  & \color{gray}{0.772}   & \color{gray}{0.932}    \\ 
\bottomrule
\end{tabular}}
\caption{\textbf{Long-term action anticipation on Ego4D.} Edit distance values for forecasting horizon of $Z=5$ and $Z=20$ actions are shown on v1 validation set.} 
\squeezeup
\label{tab:long_term_results}
\end{table}


\noindent \textbf{Smaller LLMs benefit from implicit representation of visual information for short to medium-range visual history.} In contrast to LTA however, VCLMs show competitive performance in the VPA task requiring grounding of medium visual history (Table~\ref{tab:vpa}). 
Specifically, VCLM 13B outperforms the Socratic 13B model by a large margin (mAcc: $21.2\rightarrow{25.5}$ and mIoU: $37.4\rightarrow45.5$ for $Z=4$) in VPA on CrossTask (Table~\ref{tab:vpa}). In Appendix~\ref{vpa_scale}, we show that this trend is consistent for 7B LLMs.
However, this performance gap between the VCLMs and Socratic models disappears for 70B LLMs in VPA. Thus, the \change{implicit vision representation may capture signals that help in forecasting, especially when using smaller LLMs (7B, 13B) with} limited reasoning and planning capabilities. However, such implicit information may not be essential for larger LLMs, which may be able to plan well with coarse-level grounding. 



\noindent \textbf{Larger language models lead to better planning with limited, unstructured information from the visual history.} Akin to many existing works~\cite{kaplan2020scaling}, we find that scaling laws hold for our video-based planning tasks. Overall action prediction performance improves in both LTA and VPA for both categories of models as LLM size increases from 13B to 70B (Tables~\ref{tab:long_term_results},\ref{tab:vpa}). 
Furthermore, in comparison to existing SOTA approaches such as AntGPT~\cite{zhao2023antgpt} (7B/13B) and Palm~\cite{huang2023palm} (1.5B/7B) in Table \ref{tab:long_term_results}, our Socratic model (70B) achieves better performance with limited and unstructured information from visual history. Specifically, AntGPT uses Llama2 13B to infer the overall goal of actions in visual history and leverages it with Chain of Thought (COT) prompting \cite{wei2022chain} to predict future actions. Palm uses an additional action recognition model -- EgoVLP \cite{lin2022egocentric} to extract actions present in the visual history and uses them along with the narrations to represent the visual history. Instead, our models only use narrations to represent the visual history.


\begin{table}[t]
\centering
\setlength{\tabcolsep}{3pt}
\resizebox{1\columnwidth}{!}{%
\begin{tabular}{@{}ccrrrrrrr@{}}
\toprule
  \multirow{2}{*}{\textbf{Model}} &
  \textbf{Supervised} &
  $Z=1$ &
  \multicolumn{3}{c}{$Z=3$} &
  \multicolumn{3}{c}{$Z=4$} \\ \cmidrule(l){3-9} 
 &
  \textbf{Samples} &
  mAcc &
  SR &
  mAcc &
  mIOU &
  SR &
  mAcc &
  mIOU \\ \midrule
  {Random} &
  &
  0.9   &
  0.0   &
  0.9   &
  1.5   &
  0.0   &
  0.9   &
  1.9  \\
  Socratic 13B &
  8 &
    22.8 &
     5.6 &
     22.2 &
     35.6 &
     3.0 &
    21.2 &
    37.4 \\
  VCLM 13B &
  8 &
     27.2 &
      6.9 &
     25.2 &
     41.7 & 
     4.3  &
     25.5 &
     45.5 \\
  Socratic 70B &
  8 &
    28.1 &
     \textbf{9.1} &
     26.6 &
     \textbf{43.6} &
     5.5 &
     25.5 &
    45.7 \\
  VCLM 70B &
  8 &
    28.1  &
      8.9 &
    \textbf{26.9} &
     43.4 &
      \textbf{6.1} &
     \textbf{26.8} &
     \textbf{46.9}  \\
     \hline
  \color{gray}{DDN (supervised) \cite{chang2020procedure, patel2023pretrained}} &
  \color{gray}{1756} &
  \color{gray}{33.4} &
  \color{gray}{6.8} &
  \color{gray}{25.8} &
  \color{gray}{35.2} &
  \color{gray}{3.6} &
  \color{gray}{24.1} &
  \color{gray}{37.0} \\ 
  \color{gray}{Ego4D (supervised) \cite{ego4d}} &
  \color{gray}{1756} &
  \color{gray}{26.9} &
  \color{gray}{2.4} &
  \color{gray}{24.0} &
  \color{gray}{35.2} &
  \color{gray}{1.2} &
  \color{gray}{21.7} &
  \color{gray}{36.8} \\ 
  \color{gray}{VLaMP (supervised) \cite{patel2023pretrained}} &
  \color{gray}{1756} &
  \color{gray}{50.6} &
  \color{gray}{10.3} &
  \color{gray}{35.3} &
  \color{gray}{44.0} &
  \color{gray}{4.4} &
  \color{gray}{31.7} &
  \color{gray}{43.4} \\ 
  \bottomrule
\end{tabular}%
}

\caption{\textbf{Short and medium horizon visual planning (VPA) on CrossTask.} Mean accuracy, mean IoU, and Success Rate (SR) percentages are shown for short, $Z=1$, and medium, $Z=3,4$, horizons. We follow the prior work on VPA~\cite{patel2023pretrained} for generating narrations while Internvideo~\cite{wang2022internvideo} is used as the video encoder for VCLM as in LTA.}
\squeezeup
\vspace{.3cm}
\label{tab:vpa}
\end{table}


\noindent \textbf{LLMs reduce the need for supervision in long-horizon planning.} Few-shot Socratic and VCLMs achieve performance comparable to the fully supervised models on LTA (Table \ref{tab:long_term_results}) as well as on VPA for longer-horizon predictions $Z=4$ (Table~\ref{tab:vpa}). Notably, as highlighted in Tab.~\ref{tab:vpa}, such competitive performance is achieved while using only a fraction of supervision (8 examples in few-shot models vs. 1756 examples for finetuning VLaMP). Overall, the ability of the models to capture a broader but probable distribution of action cooccurrences and dependencies given coarse context may be more important for longer horizon predictions. Consequently, the common sense of LLMs shines in these settings leading to competitive performance despite no task-specific finetuning.



\vspace{-.1cm}
\section{User-in-the-loop Evaluation}
\label{sec:online}
\vspace{-.1cm}
Our benchmarking experiments on LTA and VPA highlight the strengths and weaknesses of VCLMs and Socratic approaches to predict future actions based on video history. However, it is unclear how these actions would manifest with a user in the loop and whether these actions -- when executed -- would successfully complete real-world activities. To that end, we conduct an online evaluation of VCLMs and Socratic approaches in real-world assistive scenarios. We recruit 18 participants to perform multi-step cooking activities while wearing an egocentric observation device called Aria~\cite{project_aria} and following assistance from one of these models. We measure the true activity completion success rate, which is difficult to measure offline, and the correctness of recommended actions using mean IoU.


\begin{figure*}[t]
    \centering
    \includegraphics[width=.85\textwidth]{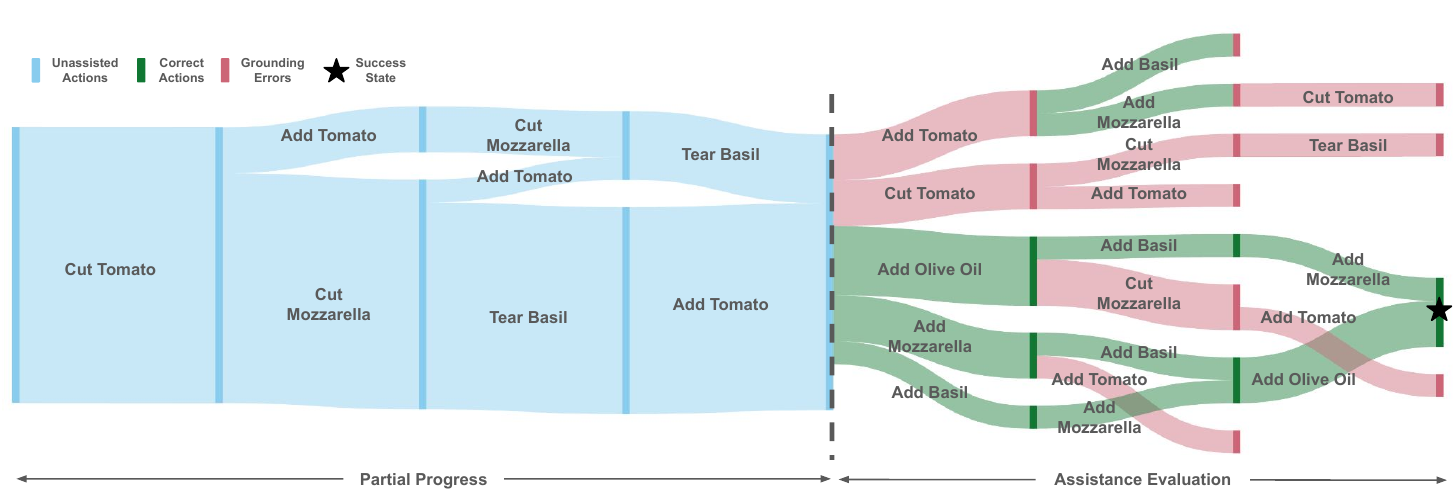}
    \caption{\textbf{Overview of the flow of steps for the activity of making caprese salad}. Unassisted actions performed by the participants in the \emph{partial progress} phase of the activity followed by multimodal LLM assisted activity execution is shown.}    
    \label{fig:online_task_flow}
    \vspace{-0.5cm}
\end{figure*}
\vspace{-.1cm}
\subsection{Study Design}
\vspace{-.1cm}
\noindent \textbf{Multi-step activities.} We choose three cooking activities for our study: 1) espresso latte, ~2)~caprese salad, and~3)~BLT (bacon lettuce and tomato) sandwich. These activities consist of a variety of ingredients, including meat, vegetables, breads, and liquids, and require different types of actions, including pouring, chopping, spreading, and plating. Furthermore, these activities also have some ordering constraints among the steps. For instance, the milk needs to be frothed before pouring into the espresso for making latte and the BLT ingredients need to be stacked on the bread before closing the sandwich. Lastly, we account for the ease of doing these activities in an office kitchen, which required omitting really long activities or activities using a stove, oven, etc.

\noindent \textbf{Study protocol.} 
Each participant performs two of the three aforementioned activities instructed by either a VCLM or a Socratic model. We use Latin-square counterbalancing for the ordering of multimodal LLM that offers them assistance as well as the type of activity that they perform across participants to reduce learning effect~\cite{john1998statistical}.

Each activity entails a script that a participant can follow. Full scripts of the three activities can be found in Appendix~\ref{sup:task_scripts}. Each activity script is split into two phases --~1)~\emph{Partial progress:} In the initial phase of the activity, participants are asked to make partial progress in the activity by completing a set of steps in any feasible order. For example, in the caprese-making activity, participants are instructed to slice tomatoes and mozzarella, tear basil, and place tomato slices on the plate. They are also free to slice varying amounts of these ingredients in whatever manner they like e.g., small vs. large slices.~2)~\emph{Assistance evaluation:} In this phase of the activity, the multimodal LLM assistant engages and guides the participant through completing the remaining steps. Participants iteratively request the next task step from the assistant and then execute what the assistant asks them to do to the best of their ability. Figure~\ref{fig:online_task_flow} gives an overview of these phases for the caprese-making activity. Participants may skip recommended actions from the assistant that are infeasible, irrelevant, or already completed. Actions are skipped solely at the participant's discretion. The activity episode is considered complete if the assistant returns a ``done" step (for example, asking the participant to serve their dish), if the participant chooses to skip 3 instructions in a row, or after the participant executes $n + 2$ actions, where $n$ is the number of steps in the evaluation section of the script. We allow $n+2$ actions so as to account for multiple successful action sequences, including ones with optional steps (Fig.~\ref{fig:latte_task_flow}).

\noindent \textbf{Evaluation protocol.}
At the end of each activity episode, participants are asked to evaluate whether the food item they produced with the assistant's help is consistent with their idea of the food item they were supposed to prepare in the activity. If they are unfamiliar with the food item, they may conduct an internet search first to determine the characteristics of the item. Independently, the study administrator also evaluates whether the participant's product matches the food description. We consider an activity episode to be successful if both the participant and the administrator rate the episode as successful. \change{Since the activity can be accomplished in multiple ways and with optional steps, our approach of using two human ratings for estimating activity success ensures a conservative and robust measurement}. We also record individual actions recommended by the assistant and whether they were skipped, executed, or infeasible to compute the mean IoU and analyze the types of errors.

\vspace{-.1cm}
\subsection{Real-world Deployment of Multimodal LLMs}
\label{sec:online_system}
\vspace{-.1cm}
We frame the multimodal LLM assistance in our online study after the VPA task~\cite{patel2023pretrained} used in our offline benchmarking experiments (Sec.~\ref{sec:offline_benchmarks}), since we believe its definition is closest to vision-based assistants. Following VPA, the VCLMs and Socratic models are given an untrimmed and unsegmented egocentric video stream from the \emph{partial progress} phase of the activity. This usually corresponds to 3-5 high-level actions on average. We also provide the models with a natural language goal describing the activity, as in VPA. The models are then prompted to iteratively output single-step action predictions to guide the user through the remaining 2-3 steps in the \emph{assistance evaluation} phase of each activity. Akin to our offline benchmarking, the models in our study did not have access to the activity scripts, nor had they seen the kitchen environment where the experiments were conducted. We use the same retrieval-based few-shot prompting strategy as in our offline experiments to obtain predictions from these models.  

\noindent \textbf{Model modifications for offline $\rightarrow$ online.} To keep inference times short during the study, we only use 13B versions of our VCLMs and Socratic models. However, direct deployment of these offline models on the online video stream from Aria does not work out of the box. The visual history accumulated in \emph{partial progress} phase of activities in the study can consist of up to 1500 frames, corresponding to 2+ minutes of video, and are akin to the visual histories in LTA. However, unlike LTA, where ground-truth segmentation of these long video histories is available to generate text-based representations and vision embeddings for Socratic and VCLM, respectively, the video history from Aria is unsegmented. To support the grounding of long, unsegmented visual history in Socratic and VCLM, we make two main modifications. First, we perform segmentation. However, the addition of yet another model, e.g., a video segmentation model in our processing pipeline, could increase computation time and lead to interaction delays in our user-in-the-loop setup. Therefore, we uniformly segment the Aria stream into clips before passing them to our narration model -- LaViLa. To compensate for unrelated and repeated narrations emerging from the uniformly segmented stream, we generate and cluster multiple narrations per segment as well as across segments based on the semantic similarity of narrations. Despite such stream segmentation and narration clustering, we find that the narrations tend to be extremely low-level, which leads to a very long narration history -- ultimately exceeding the context window of our multimodal LLMs. Hence, our second modification entails the addition of a goal-conditioned summarization step to produce the final set of narrations for encoding the long visual history in online settings. Appendix~\ref{sup_goal_sum} provides additional details about these modifications. Lastly, unlike offline benchmarking experiments where we match the open-set model outputs to a closed-set of actions (Sec~\ref{sec:offline_benchmarks}), we directly use the open-set output for easier interactions with the user in the loop. No other modifications were made to the models for online deployment. Table \ref{tab:vpa_online} in the appendix shows these online-modified models perform similarly on the VPA task.



\noindent \textbf{System setup.} We obtain the RGB video stream from Aria donned by our participants at 10 frames per second over wifi to a local machine. The frames are then center-cropped and downsampled in resolution ($1400\times1400 \rightarrow 288\times384$) to match the resolution of the LaViLa encoder. These frames are then sent to a remote server, which hosts the multimodal LLMs. The step suggestions returned by the models are parsed and communicated to the user via text-to-speech over wireless earbuds. The LaViLa narrator model runs on two-second clips of video and outputs 10 narrations per clip pre-clustering. The summarization step runs over the entire clustered narration history before every prediction step.

\vspace{-.1cm}
\subsection{Quantitative Results}
\vspace{-.1cm}
\noindent \textbf{The Socratic appraoch outperforms VCLMs at user-in-the-loop activity assistance.} Table~\ref{tab:online_success} shows the results of our online study. Akin to our offline experiments that showed the Socratic approach outperforming VCLMs in LTA (Table~\ref{tab:long_term_results}) and demonstrate competitive performance with VCLMs in VPA (Table~\ref{tab:vpa}), we find that the Socratic approach enables higher activity completion success rate as well as mIoU across the 18 participants and 3 activities in our online study. Note that our online Socratic model does not leverage finetuned video narration models like our offline experiments. Nevertheless, it exhibits superior performance. Despite the low overall success rate of both models, in 40\% of successful trials, the Socratic model enabled a user to complete a task they had not previously done. 

\begin{table}[t]
\centering
\scriptsize
\resizebox{.99\columnwidth}{!}{%
\begin{tabular}{lccc}
\toprule
\multicolumn{1}{l}{\textbf{Method}} & \multicolumn{1}{c}{\textbf{Visual History}} & \textbf{Success Rate} & \textbf{mIoU}   \\ \midrule
\multicolumn{1}{l}{Socratic 13B} & \multicolumn{1}{c}{T}   & \textbf{27.8} & \textbf{30.4}\\

\multicolumn{1}{l}{VCLM 13B} & \multicolumn{1}{c}{V+T}  &   16.7  & 23.0 \\

\midrule

\multicolumn{1}{l}{\color{gray}Socratic 13B (Offline)} & \multicolumn{1}{c}{\color{gray}T}   & \color{gray}- & \color{gray}40.3\\

\multicolumn{1}{l}{\color{gray}VCLM 13B (Offline)} & \multicolumn{1}{c}{\color{gray}V+T}   & \color{gray}- & \textbf{\color{gray}44.0}\\
\bottomrule
\end{tabular}
}
\caption{\textbf{Activity completion success rate and mean IoU metrics in percentage for the online study across all participants and activities.} Socratic models outperform VCLMs in online settings. We also rerun the models offline on the videos collected from our study to compare offline vs. online metrics. While success rate cannot be compared across offline and online settings, we find that mIoU also doesn't translate across the two settings. 
}
\label{tab:online_success}
\squeezeup
\end{table}

\begin{figure*}[t]
    \centering
    \includegraphics[width=.85\textwidth]{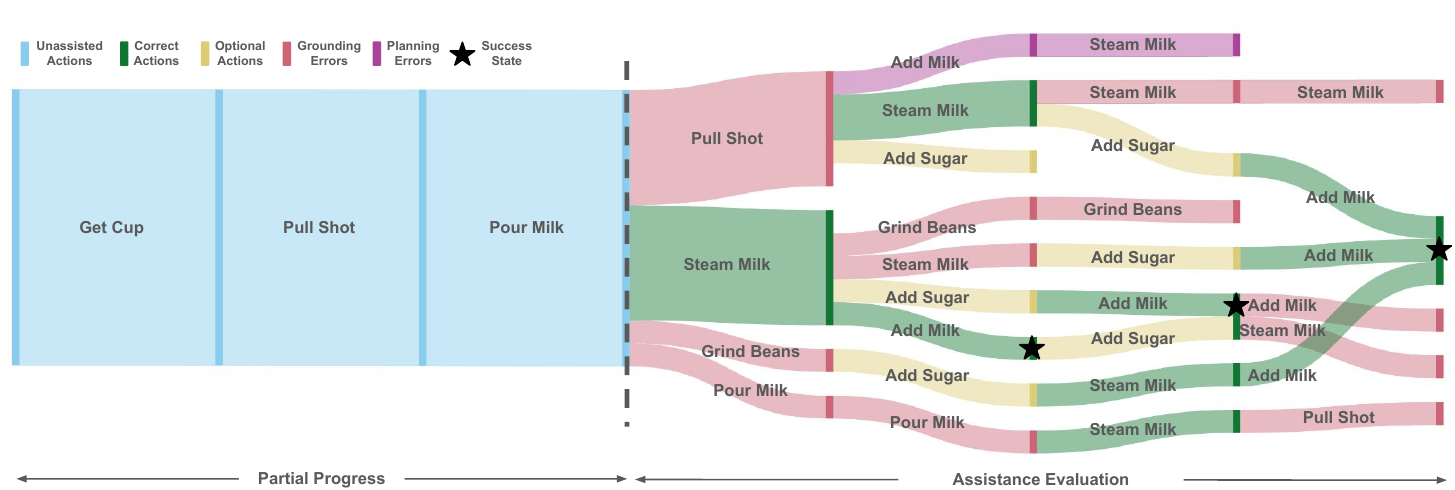}
    \caption{\textbf{Various error modes of multimodal LLMs in the latte activity}. Models fail to ground the steps that are already completed, recommend steps with incorrect ordering (planning error), or fail to recognize that the activity is complete.}
    \vspace{-0.65cm}
    \label{fig:latte_task_flow}
\end{figure*}

\noindent \textbf{Offline metrics do not capture online performance.} The success rate of activity completion cannot be truly measured in offline datasets. However, it is unclear whether other metrics used for evaluating video-based action anticipation and planning such as mIoU and edit distance (Sec.~\ref{sec:offline_benchmarks}) translate from offline settings to online settings. Despite being small scale as compared to datasets in our offline benchmarking experiments, the video data from our online experiments enable a unique opportunity to compare these metrics head-on in both online and offline settings. 
To this end, we rerun Socratic and VCLM offline in the videos from our study. Specifically, we provide the models with videos from the \emph{partial progress} phase of the activities along with the activity goal, e.g., ``make Caprese salad with mozzarella, tomato, basil, olive oil" following the VPA task (Sec.~\ref{sec:offline_benchmarks}). The models are then prompted with few-shot examples, following our prompt setup from offline experiments, to predict $n + 2$ steps. Here, $n$ is the expected number of steps remaining in the activity from the \emph{assistance evaluation} phase of the activity.
We observe higher mean IoU rates for both models when run offline compared to their mIoU when they provided user-in-the-loop assistance online. Furthermore, VCLM outperforms Socratic in offline mIoU. However, it lags behind in both online mIoU and real-world success rates, indicating that offline mIoU may be an unreliable predictor of real-world performance. The gap between offline and online mIoU may partly be attributed to the single multi-step prediction of all the remaining actions in the offline setting versus iterative single-step predictions, i.e., with replanning in the online setting. The iterative single-step predictions are more likely to make repeat suggestions, often due to grounding errors, which lead to a lower intersection between suggested steps and ground-truth steps. 




\vspace{-.1cm}
\subsection{Qualitative Analysis of Model Errors}
\vspace{-.1cm}
\noindent \textbf{Grounding errors, planning errors, and failure to detect activity end/success are the main failure modes.} We also evaluate cases where a participant skipped actions recommended by the assistant. Recall that participants could skip action recommendations that were redundant, infeasible, or irrelevant. Appendix~\ref{sup:error_analysis} shows a detailed analysis and breakdown of these reasons for skipped actions by the participants across the VCLM and Socratic models per activity. Such analysis enabled us to identify three main error modes -- grounding errors, planning errors, and failure to detect activity end/success. Figure~\ref{fig:latte_task_flow} shows these error modes for the espresso latte activity across all participants. Redundant skipped actions often correlate to grounding mistakes, where the models fail to recognize a step that has already been completed. Infeasible skipped steps also often correlate with grounding errors. Here, the models may suggest something that works for a different variation of the activity. For example, grinding coffee might work for a different version of a latte-making activity, but participants used an automated espresso machine in our study. These grounding errors are often more subtle than grounding errors from redundant steps. Finally, irrelevant skipped steps often correlate to planning errors, where the models suggest a step that is not part of the activity. We also find that in $50\%$ of the successful activity episodes, the models fail to recognize when an activity is completed. 


\noindent\textbf{Offline metrics don't capture error modes.} 
The failure of models to detect activity end/success state does not affect the success rate metric, which would count such activity episodes as successful. However, it does lower mIoU scores due to redundant suggested actions. The overview of the participant steps for making a latte in Fig.~\ref{fig:latte_task_flow} succinctly highlights the known issues with offline metrics. Specifically, mIoU as a permissive metric, would consider adding milk before steaming, a planning error, as a success. Conversely, offline success rate, being a restrictive metric, would discount 4 of the 5 present paths to success for making lattes as failures. Furthermore, mIoU and edit distance metrics do not capture optional actions sometimes suggested by such models that do not affect activity success e.g., adding sugar.


\noindent \textbf{Grounding errors are the dominant mode of failure for models online.} The bulk of the errors both models exhibit pertain to grounding. In particular, past participant actions are either not captured by the narrations or visual embedding of their activity history or are present in the long history but not attended to by the models during prediction, leading to grounding errors. We find that 63\% of the skipped action recommendations are due to redundant action suggestions emerging from erroneous grounding (Appendix~\ref{sup:error_analysis}). The distribution of skipped actions is consistent across both models and indicates that recognizing previously completed actions in an activity is an ongoing challenge for these models. In contrast, both models make fewer planning errors, i.e., they suggest fewer irrelevant actions or actions with incorrect orderings. 
Overall, our analyses of skipped actions and errors in the study indicate that the primary challenge with visual assistants still lies in reasoning about activity progress and activity success/failure via grounding -- more so than task knowledge or planning.

\vspace{-.2cm}

\vspace{-.1cm}
\section{Conclusion}
\vspace{-.1cm}
We evaluate the two predominant multimodal LLM-based approaches: Vision Conditioned Language Models (VCLM) and Socratic Models for vision-based activity assistance through two video-based action anticipation benchmarks on \emph{offline} datasets and a real-world \emph{online} study with 18 participants. To the best of our knowledge, our online evaluation is the first of its kind for multimodal LLMs towards real-world activity assistance systems. Our experiments show the Socratic approach is better equipped to capture coarse visual details across a long visual history. Current VCLMs can capture more fine-grained details but only for short visual history. Encoding long videos and aligning long videos with text tokens as needed by VCLMs would thus be rich avenues for future work. In the interim, Socratic models demonstrate competitive behaviors on video-based action anticipation and planning tasks spanning short to long visual history both offline and online. 



Our work sets important directions for future research on multimodal LLMs as vision-based assistants. Our online study highlights how grounding is the largest source of errors for these types of models. Grounding at different granularities remains an open problem, which, when improved, will greatly enhance activity assistance systems. Furthermore, we show how offline metrics do not provide a good indication of performance in online settings, demonstrating the importance of real-world evaluation of models for assistive scenarios. 




%
%
\bibliographystyle{splncs04}
\bibliography{PaperForReview}

\clearpage

\noindent \textbf{This appendix is organized as follows:} \\
\noindent 7. Offline and Online Experiment Details. 

\noindent 8. Prompt templates for LTA and VPA. 

\noindent 9. Ablations related to visual history representation. 

\noindent 10. Tabular Comparison of VCLM and Socratic Models

\noindent 11. Activity scripts for online evaluation. 

\noindent 12. Model Error Analysis in Online Evaluation. 

\noindent 13. Study Activity Visualization.

\noindent 14. Participant Data Collection Practices



\section{Offline and Online Experiment Details}
\label{sup_ex_details}

\subsection{Offline Benchmark Tasks}
We evaluate VCLM and Socratic models on two existing video-based forecasting benchmarks -- Long Term Action Anticipation (LTA)~\cite{ego4d} and Visual Planning for Assistance (VPA)~\cite{patel2023pretrained} using offline datasets -- Ego4D~\cite{ego4d} and CrossTask~\cite{zhukov2019cross} respectively  (Sec.~\ref{sec:offline_benchmarks}). Here, we provide a detailed overview of the datasets and our experimental setup for each of these tasks. 



\noindent\textbf{Ego4D-LTA} \cite{ego4d}:  Ego4D consists of 3,670 hours of video footage of everyday activities, with 53 different scenarios. Out of this, we use the LTA (forecasting) subset, which entails 116 hours. This subset contains 1723 clips that cover an action space of 115 verbs and 478 nouns. We use the standard train and validation splits proposed by Ego4D \cite{ego4d} for our evaluation. In the LTA task, given 8 video segments from a video clip as input, the models must predict the 20 future actions in the form of verb, noun, and verb + noun, in correct order. Edit distance between the predicted sequence of actions and the ground truth action sequence in the video clip is used as a metric for evaluation following Ego4D~\cite{ego4d}. 

\noindent\textbf{CrossTask-VPA} \cite{zhukov2019cross}: 
CrossTask consists of 2.7K instructional videos for 18 different tasks from multiple domains, covering 374 hours of footage. Some of the action classes were shared among different tasks, with a total of 118 actions. Each video consists of an average 7.6 action steps. We follow \cite{patel2023pretrained} to construct a train split with 1,564 videos and a test split with 752 videos. We extract multiple test samples from each test video for VPA -- specifically, given an annotated video consisting of K steps, we generate $K - Z$ samples,  leaving at least $Z = 3, 4$ steps to predict in the future. This leads to a dataset of 4123 test samples for our evaluation. Our VPA task definition also follows~\cite{patel2023pretrained} -- given an untrimmed video and a goal of the activity/task in the video described in natural language as input, the models must predict the up to 4 future actions in the form of verb+noun, in correct order. Evaluation compares the predicted action sequence with the ground truth actions in the video using mIoU, per step accuracy, and success rate metrics (Sec.~\ref{sec:offline_benchmarks}).

\subsection{Model Modifications for Online Evaluation}
\label{sup_goal_sum}
\noindent \textbf{Goal-Conditioned Summarization.}
Sec.~\ref{sec:online_system} provides an overview of modifications for our multimodal LLMs to enable online evaluation. The online settings entail noisy stream of redundant video frames leading to long narration history. To handle such long narration histories in a robust manner, one of the biggest changes we make in these models is goal-conditioned summarization.
This greatly reduces the number of tokens in the input and allows the language model to attend to a longer narration histories more robustly while still leveraging few-shot examples. The summarization is performed by LLama2-13B Chat using the following prompt:

\lstset{frame=tb,
  aboveskip=3mm,
  belowskip=3mm,
  showstringspaces=false,
  columns=flexible,
  basicstyle={\small\ttfamily},
   commentstyle=\color{gray},
   stringstyle=\color{black},
  numbers=none,
  breaklines=true,
  breakatwhitespace=true,
  tabsize=2
}
\begin{lstlisting}[language=Go]
A person is currently attempting to 
[goal]. Their task is in progress and 
their goal is not yet complete. The 
following are low level narrations of 
their actions.

[narration history]

Please summarize these into a smaller
set of high-level narrations. Focus on
narrations that are relevant to the 
goal and do not include irrelevant 
narrations in your high-level summary.
Begin every high-level narration with 
the text, 'A person ':
1. A Person 
\end{lstlisting}

The output from the LLM is parsed by only keeping lines starting with numbers. The helps remove any conversation or filler language in the response. This summarized history often contains 5-20 high-level narrations and is used by the LLM to perform prediction. Few-shot examples are also summarized offline.

\noindent\textbf{Goal-Generation for Few-shot Examples.}
We evaluate the utility of goal conditioning in online experiments, akin to our offline experiments (Sec.~\ref{sec:goalcond}). To that end, our pilot studies show that goal-conditioned prediction performs much better than prediction without goals in the online setting. Specifically, we find that goals help the LLM identify which parts of noisy input video stream and narrations are relevant to completing the activity. In order to ensure our few-shot examples to the multimodal LLMs, which are obtained from Ego4D, are appropriately goal conditioned for online experiments, we need to annotate these with goal information. Since, goal information is not available in Ego4D, we again resort to LLMs for obtaining pseudo goal labels for these videos. Specifically, we use Llama2-70B chat with the following prompt to generate goals for Ego4D LTA training set videos:

\lstset{frame=tb,
  aboveskip=3mm,
  belowskip=3mm,
  showstringspaces=false,
  columns=flexible,
  basicstyle={\small\ttfamily},
   commentstyle=\color{gray},
   stringstyle=\color{black},
  numbers=none,
  breaklines=true,
  breakatwhitespace=true,
  tabsize=2
}
\begin{lstlisting}[language=php]
The user took these physical actions:
[Narration History]

What are the top 3 goals of the user?

Respond only in JSON that satisfies the Response type:
type ResponseList = [Response_1, Response_2, ..., Response_3] 
type Response = {
user_goal: str;
confidence: float;
explanation: str;
} 
Provide {user_goal} in the format of 'They wanted to {user_goal}', the {confidence} of the goal given the context (on a scale from 0 to 1), and a terse {explanation} of the given goal and its confidence. 
\end{lstlisting}

where \verb|[Narration History]| is the full narration history for the clip generated by LaViLa. We parse the output text as a JSON and select the goal with the highest confidence.

\lstset{frame=tb,
  aboveskip=3mm,
  belowskip=3mm,
  showstringspaces=false,
  columns=flexible,
  basicstyle={\small\ttfamily},
   commentstyle=\color{gray},
  numbers=none,
  breaklines=true,
   stringstyle=\color{mauve},
  breakatwhitespace=true,
  tabsize=2
}

\noindent\textbf{Performance Comparison Between Offline and Online Models.} We also evaluate our online-modified models on the VPA task to ensure our modifications do not drastically alter performance. Table \ref{tab:vpa_online} shows the performance difference between online and offline models is relatively minimal for $Z=1$ despite the online models replacing the explicit segmentation model with uniform segmentation for faster inference. Note that in the online setting, the models provide only the next action to the user, wait for the user to execute that action, and then replan, which is a prediction horizon of $Z=1$. Prior work has shown that poor segmentation can reduce performance on the VPA task by up to 50\% \cite{patel2023pretrained}. Our online modified models see a maximum 14\% drop in performance, which indicates our online modifications (clustering and summarization) help mitigate the performance drop from our simplified segmentation.
\begin{table}[h]
\vspace{-.37cm}
\resizebox{1\columnwidth}{!}{%
\begin{tabular}{crrrrrrr}
\hline
\multirow{2}{*}{Model} & \multicolumn{1}{c}{Z=1}  & \multicolumn{3}{c}{Z=3}                                                      & \multicolumn{3}{c}{Z=4}                                                      \\ \cline{2-8} 
                       & \multicolumn{1}{l}{mAcc} & \multicolumn{1}{l}{SR} & \multicolumn{1}{l}{mAcc} & \multicolumn{1}{l}{mIOU} & \multicolumn{1}{l}{SR} & \multicolumn{1}{l}{mAcc} & \multicolumn{1}{l}{mIOU} \\ \hline
Socratic Online           & 22.5                     & 2.3                    & 17.9                     & 29.8                     & 1.1                    & 17.8                     & 34.7                  \\
VCLM Online               & 23.3                     & 4.3                    & 18.5                     & 33.2                    & 1.8                    & 18.9                     & 41.3                  \\
Socratic Offline           & 22.8                     & 5.6                    & 22.2                     & 35.6                     & 3.0                    & 21.2                     & 37.4                   \\
VCLM Offline               & 27.2                     & 6.9                    & 25.2                     & 41.7                     & 4.3                    & 25.5                     & 45.5                   \\ \hline
\end{tabular}
}
\caption{\textbf{Comparison between online and offline models on the VPA task.} Note that the online models suggest the next step to the user, wait for the user to execute that step, and then replan ($Z=1$).}
\label{tab:vpa_online}
\vspace{-.45cm}
\end{table}
\section{Prompt Templates for LTA and VPA}
\label{sup:prompt_template}

Detailed prompt templates for our offline benchmark tasks LTA and VPA as shown in figures \ref{fig:prompt_lta} and \ref{fig:prompt_vpa}. The prompt for LTA (Fig.~\ref{fig:prompt_lta}) consists of \textit{examples} text narration sequences pertaining to the full video from 8 videos of the training set and the \textit{visual history} of 8 segments from the current video. The narrations are from the LaViLa narration model \cite{zhao2023lavila}. 
Likewise, the prompt for VPA (Fig.~\ref{fig:prompt_vpa}) includes \textit{examples} of full action sequences consisting of ground truth (GT) action labels for 8 videos from the training set and the \textit{visual history} of the current video, which entails actions predicted following previous work~\cite{patel2023pretrained} noted as [predicted action].




\begin{figure}[htb]
\begin{tabular}{l}

\begin{lstlisting}
"Task description"
#Prompt example *8 from training set:
    1. [narration]  
    2. [narration]     
    ... 
    N. [narration]
 

#Visual history from current video:
    T-8. [narration] 
    ...
    T-1. [narration]
    T.    
\end{lstlisting}
 \end{tabular}
\caption{\textbf{Prompt template for Ego4D LTA}. We set $N$ to be the total number of actions in the video and $T$ to be the starting action index that we want to predict in the current video.}
\label{fig:prompt_lta}
\end{figure}

\begin{figure}[htb]
\begin{tabular}{l}

\begin{lstlisting}[language=Python]
"Task description"
#Prompt example *8 from training set:
    Goal: [CrossTask Task Title]
    1. [GT action]  
    2. [GT action]     
    ... 
    N. [GT action]
 

#Visual history from current video:
    Goal: [CrossTask Task Title]
    1. [predicted action]  
    ...
    T-1. [predicted action]
    T.    
    
\end{lstlisting}
 \end{tabular}
\caption{\textbf{Prompt template for CrossTask VPA}. We use the video's task title from CrossTask as goal description for VPA and append it in the front of the action sequence in our prompts for VPA. We use predicted actions following previous work \cite{patel2023pretrained} to construct the visual history of the current video.
$N$ and $T$ follow the same design as in LTA.}
\label{fig:prompt_vpa}
\end{figure}

\section{Ablations on Visual History Representation}\label{sub_additional_ex}

\subsection{Evaluation of the benefit from implicit representation of visual information for smaller LLMs across different LLM sizes.} \label{vpa_scale}
Table \ref{tab:vpa_scale2} above shows that mAcc gap in VPA task for 7B models with and without visual conditioning at Z = 1,3,4 is 5.6\%, 3.8\%, and 3.2\% respectively. The mAcc gap for 13B models at Z = 1,3,4 is 4.4\%, 3\%, and 4.3\% and for 70B models is 0\%, .3\%, and 1.3\% respectively as in Table \ref{tab:vpa}. Implicit visual representation aids smaller LLMs across model sizes.


\begin{table}[t]
\centering
\setlength{\tabcolsep}{3pt}
\resizebox{\columnwidth}{!}{%
\begin{tabular}{@{}ccrrrrrrr@{}}
\toprule
  \multirow{2}{*}{\textbf{Model}} &
  \textbf{Supervised} &
  $Z=1$ &
  \multicolumn{3}{c}{$Z=3$} &
  \multicolumn{3}{c}{$Z=4$} \\ \cmidrule(l){3-9} 
 &
  \textbf{Samples} &
  mAcc &
  SR &
  mAcc &
  mIOU &
  SR &
  mAcc &
  mIOU \\ \midrule
  {Random} &
  &
  0.9   &
  0.0   &
  0.9   &
  1.5   &
  0.0   &
  0.9   &
  1.9  \\
  Socratic 7B &
  8 &
    22.3 &
     4.3 &
     21.0 &
     33.3 &
     2.6 &
    20.8 &
    36.2 \\
  VCLM 7B &
  8 &
     27.9 &
      6.8 &
     24.8 &
     41.7 & 
     4.1  &
     24.0 &
     45.0 \\
  Socratic 13B &
  8 &
    22.8 &
     5.6 &
     22.2 &
     35.6 &
     3.0 &
    21.2 &
    37.4 \\
  VCLM 13B &
  8 &
     27.2 &
      6.9 &
     25.2 &
     41.7 & 
     4.3  &
     25.5 &
     45.5 \\
  Socratic 70B &
  8 &
    28.1 &
     \textbf{9.1} &
     26.6 &
     \textbf{43.6} &
     5.5 &
     25.5 &
    45.7 \\
  VCLM 70B &
  8 &
    28.1  &
      8.9 &
    \textbf{26.9} &
     43.4 &
      \textbf{6.1} &
     \textbf{26.8} &
     \textbf{46.9}  \\
  \bottomrule
\end{tabular}%
}
\vspace{.2cm}
\caption{\textbf{Varying LLM size in visual planning (VPA) on CrossTask.} Mean accuracy, mean IoU, and Success Rate (SR) percentages are shown for short $Z=1$ and medium $Z=3,4$ horizons.  We use predicted actions from a finetuned segmentation model following previous work \cite{patel2023pretrained} to construct the visual history of the current video.}
\vspace{-0.3cm}
\label{tab:vpa_scale2}
\end{table}

\subsection{Task-relevant information from visual history}
Different aspects of the visual history can be extracted and represented in text for VCLMs and Socratic models. It is unclear what aspects should be extracted to enable efficient forecasting in such models. To that end, we compare different modes of task-relevant information for Socratic multimodal LLMs on the Ego4D LTA task. Each of these modes of information can be obtained from different pre-trained vision-language models. Specifically, we consider information on objects, actions, and narrations describing activities in the video as the three relevant information modes. 

We obtain object descriptions using Detic~\cite{Zhou2022DetectingTC} with the Ego4 LTA noun set as a custom vocabulary, recognized actions using the LaViLa dual encoder with the Ego4D LTA closed-set of actions, and open-set narrations using the LaViLa narration model~\cite{zhao2023lavila}. We test three settings i.e., combinations of these information modes: only narrations, narrations and objects, narrations and actions. The only narrations setting uses the same prompts as our Socratic model described in Sec~\ref{sec:offline_mm_llm}. The narrations and objects setting prepends a list of recognized objects from the input video before the narrations in the visual history. The narrations and actions setting follows the same prompting structure as Palm~\cite{huang2023palm}. All three settings use the same retrieval-based prompting approach as described in Sec~\ref{sec:experiment_setup}. Action and object prompts are generated on the LTA train set. 

\begin{table}[t]
\centering
\scriptsize
\resizebox{\columnwidth}{!}{
\begin{tabular}{lclll}
\toprule
\multirow{2}{*}{\textbf{Information Type}}               & \multirow{2}{*}{\textbf{VLM}}                        & \multicolumn{3}{c}{ED@(Z=20)↓} \\ \cmidrule(l){3-5} 
                     &                                       & Verb    & Noun    & Actions    \\ \midrule
Narrations Only      & LaViLa                        & \textbf{.731}    & .787    & \textbf{.951}       \\
Narrations + Objects & LaViLa  + Detic               & .734    & \textbf{.776}    & .952       \\
Narrations + Actions & LaViLa  + LaViLa Dual Encoder & .732    & .812    & .958       \\ \bottomrule
\end{tabular}}
\vspace{.2cm}
\caption{\textbf{Comparison of different information types/modes that can represent a video's history using Socratic models on LTA.} Edit distance values for forecasting horizon of $Z=20$ actions is shown on v1 validation set.}
\vspace{-0.2cm}
\label{tab:socratic_comparison}
\end{table}

Table~\ref{tab:socratic_comparison} shows the results for these three settings on Ego4D LTA. All models use Llama2-7B as the LLM. As seen in the table, neither adding object nor action information from the visual history noticeably improves performance on the LTA task. Following this result, we determine that object and closed-set action information is a subset of open-set narration information when it comes to visual history representation. Consequently, we use only narrations to represent visual history for both VCLM and Socratic models in all our offline and online experiments (Sec.~\ref{sec:offline_benchmarks}, \ref{sec:online}).


\subsection{Comparison of narrators for visual history}
Video history might be sufficiently represented using open-set narrations that describe the activity in the video (Tab.~\ref{tab:socratic_comparison}) for video-based forecasting tasks. To determine an appropriate video narration model for our multimodal LLMs in forecasting tasks, we compare two SOTA video narrators -- LaViLa~\cite{zhao2023lavila} and the Blip-2~\cite{blip2}. We perform this comparison using the Llama2-7B Socratic models on LTA (Table~\ref{tab:narrator_comparison}).  
Following Palm~\cite{huang2023palm}, we feed the median frame from a video segment along with the prompt ``A person is~" to the Blip-2 model for generating a narration describing the video segment. In contrast, the LaViLa narration model uses 4 evenly spaced frames. We parse the output of the narrator model by replacing references to the participant with ``A person" to ensure consistent structure. Note that, LaViLa narrator uses GPT2-XL (1.5 B parameters) as its LLM backbone while Blip-2 uses OPT 2.7B. LaViLa is also explicitly trained on Ego4D to narrate video clips~\cite{zhao2023lavila}.


\begin{table}[t]
\centering
\scriptsize
\resizebox{1\columnwidth}{!}{
\begin{tabular}{lcclll}
\toprule
\multirow{2}{*}{\textbf{Narrator}} & \multirow{2}{*}{\textbf{Input Frames}} & \multirow{2}{*}{\textbf{Narrator LM}} & \multicolumn{3}{c}{ED@(Z=20)↓} \\ \cmidrule(l){4-6}  
                &              &                & Verb    & Noun    & Actions    \\ \midrule
LaViLa  & 4            & GPT2-XL (1.5B) & \textbf{.731}    & \textbf{.787}    & \textbf{.951}       \\
Blip-2          & 1            & OPT 2.7B       & .758    & .883    & .978       \\ \bottomrule
\end{tabular}}
\vspace{.2cm}
\caption{\textbf{Comparison of narrators using Llama2-7B Socratic models on LTA.} Narrator LM represents the LLM backbone for each narrator model. Edit distance values for forecasting horizon of $Z=20$ actions is shown on v1 validation set.}
\vspace{-0.2cm}
\label{tab:narrator_comparison}
\end{table}

As seen in Table~\ref{tab:narrator_comparison}, despite its smaller language model backbone, LaViLa narrator significantly outperforms Blip-2 for capturing relevant visual history for forecasting. This is likely due to LaViLa's narration-specific and egocentric training data, as well as consumption of 4 frames from the input video rather than just 1. Based on this analysis, we use LaViLa narrator for open-set narration generation of visual history for all our experiments unless otherwise specified.

\subsection{Medium history-medium horizon forecasting in VPA without goal}
\label{sec:goalcond}
Our offline benchmark tasks -- VPA and LTA, cover the spectrum of medium to long forecasting based on medium to long visual history respectively. However, unlike LTA, the VPA task~\cite{patel2023pretrained} also uses the goal of the activity in the video, in addition to the visual history, for forecasting future actions. To understand the performance of multimodal LLMs on medium history, medium horizon forecasting problems without the availability of goal information, we conduct an ablation on VPA. Specifically, we evaluate the best performing multimodal LLM -- VCLM 70B on VPA with and without goal information (Table \ref{tab:vpa_nogoal}).
We simply remove the goal information from the VPA prompt (Fig.~\ref{fig:prompt_vpa}) for this analysis.

The results show that the information regarding the goal enables the VCLMs to make better mid-horizon forecasting predictions while slightly decreasing the accuracy of short horizon predictions. Thus, the performance of multimodal LLMs may overall drop in medium history, medium horizon forecasting problems when goal information is not available. Since the availability of goal information leads to improved performance and since such information may be easy to obtain in user-in-the-loop settings, we frame our online evaluation using VPA's task definition i.e., with inclusion of goal. 

\begin{table}[]
\centering
\setlength{\tabcolsep}{3pt}
\resizebox{1\columnwidth}{!}{%
\begin{tabular}{@{}ccrrrrrrr@{}}
\toprule
  \multirow{2}{*}{\textbf{Model}} &
  \multirow{2}{*}{\textbf{Goal}} &
  $Z=1$ &
  \multicolumn{3}{c}{$Z=3$} &
  \multicolumn{3}{c}{$Z=4$} \\ \cmidrule(l){3-9} 
 &
   &
  mAcc &
  SR &
  mAcc &
  mIOU &
  SR &
  mAcc &
  mIOU \\ \midrule
  VCLM 70B &
  No &
    \textbf{28.6} &
     8.1 &
     26.5 &
     41.8 &
     5.5 &
     26.3 &
    45.0 \\
  VCLM 70B &
  Yes &
    28.1  &
      \textbf{8.9} &
    \textbf{26.9} &
     \textbf{43.4} &
      \textbf{6.1} &
     \textbf{26.8} &
     \textbf{46.9}  \\
  \bottomrule
\end{tabular}%
}
\vspace{.2cm}
\caption{\textbf{Short and medium horizon visual planning (VPA) on CrossTask w/wo the goal.} Mean accuracy, mean IoU, and Success Rate (SR) percentages are shown for short $Z=1$ and medium $Z=3,4$ horizons. LaViLA is used as the narration model while Internvideo is used as the video encoder for VCLM as in LTA.}
\vspace{-0.3cm}
\label{tab:vpa_nogoal}
\end{table}

\subsection{Text-based history representation for VCLMs on long-history tasks}
Since current VCLMs may be capable of encoding only limited visual history, we also provide text-based representation of history as input to VCLMs for our long history-based forecasting tasks e.g., LTA. We conduct an ablation experiment to determine the contribution of such additional text-based history representation -- when used along with visual embeddings in VCLMs. Specifically, we want to determine how VCLMs without text history perform in long-history tasks. We run this ablation with a Llama2-70B chat VCLM on LTA using the following prompt:
\\
\begin{lstlisting}[language=C]
Predict the next 20 actions in the form of (verb,noun)
\end{lstlisting}
\begin{table}[t]
\centering
\scriptsize
\resizebox{\columnwidth}{!}{
\begin{tabular}{lccclll}
\toprule
\multirow{2}{*}{\textbf{Model}} &\multirow{2}{*}{\textbf{Visual History}}  & \multirow{2}{*}{\textbf{Visual Encoder}}  &   \multicolumn{3}{c}{ED@(Z=20)$\downarrow$}  \\ 
\cmidrule(l){4-6} 
 &    &  & Verb   & Noun    & Action \\ \midrule
\multicolumn{1}{l}{VCLM 70B} & V  & Internvideo &  1.000 &  1.000   &  1.000     \\ 
\multicolumn{1}{l}{VCLM 70B} & V+T  & Internvideo+LaViLa &  0.739 & 0.731   & 0.931      \\ 
\bottomrule
\end{tabular}}
\vspace{.2cm}
\caption{\textbf{Long-term action anticipation on Ego4D w/wo text history.} Edit distance values for forecasting horizon of $Z=20$ actions is shown on v1 validation set.}
\vspace{-0.2cm}
\label{tab:lta_notext}
\end{table}

The result in Table \ref{tab:lta_notext} shows that without a text-based history representation, the VCLM model fails to output meaningful predictions. Text-based history representation both provide a template for generation and help ground the VCLM to information not captured in its 8 input frames. We show examples of generated text with and without text history below.

\noindent \textbf{With text history:}
\begin{lstlisting}[language=C]
//Generated text:
16. A person drops a garlic peel in a bowl
17. A person presses a garlic clove with the knife
18. A person presses a garlic clove with a knife
19. A person drops the garlic clove in the bowl
\end{lstlisting}
\noindent \textbf{Without text history:}
\begin{lstlisting}[language=C]
//Generated text:
Pairs e.g. (drive, road) 1. Put CDs on top of magazines . 2. Paste pictures butterflies and birds near flowers . 3. Glue colored paper sequins on the bench in front of the boat . 
\end{lstlisting}
\begin{lstlisting}[language=C]
//Generated text:
pairs. 1. tv, eyes 2. jako the bird is used.
\end{lstlisting}

\section{Overview of existing VCLM and Socratic Models}
\label{sup:model_comparison}
Tables \ref{tab:socratic_models} and \ref{tab:vclm_models} show how our implemented Socratic and VCLM models compare to other models within each approach. Socratic models differ in what kinds of text the VLM's generate (actions, narrations, objects, etc), which VLMs are used, whether prediction is conditioned on an inferred goal, how prediction is performed, either by directly generating subsequent actions, or using chain of thought reasoning, and what LLM is used. We select our implemented socratic model by testing actions, narrations, and objects on the Ego4D LTA task (table \ref{tab:socratic_comparison}). We found that adding actions or objects, provided by the LaViLa encoder\cite{zhao2023lavila} and Detic\cite{Zhou2022DetectingTC} respectively, did not improve performance over open-set narrations provided by the LaViLa narrator model \cite{zhao2023lavila}. Accordingly, our representative Socratic model implementation uses only narrations as a text-based representation of visual history.

VCLM models attach a pretrained vision encoder to a pretrained LLM by mapping outputs from the vision encoder into the token embedding space of the LLM. VCLM models primarily differ by what pretrained image or video encoder is used, how aggregation across image level features is done if an image encoder is used, whether the model explicitly aligns video representations with text representations, whether the entire model is fine-tuned on an instruction dataset, and what LLM-backbone is used. Some VCLMs for video use a pretrained video encoder that samples frames from the video and processes them together. In contrast, other video VCLMs use a pretrained image encoder on a set of sampled frames and aggregate image-level features with a separate aggregation module. This aggregation module could be a form of attention, concatenation, pooling, or convolution. Prior work has indicated no clear advantage between image or video level features\cite{lin2023videollavalearningunitedvisual}. VCLM models also employ two different methods for training, alignment training and instruction tuning. In alignment training, aggregation and projection layers are trained to better map inputs from the vision encoder space to the LLM token space using a set of video-text pairs and a contrastive loss function. In instruction tuning, the entire LLM backbone is finetuned using a multimodal instruction dataset. While not every VCLM uses both of these training methods, multiple prior works have found that alignment training followed by instruction tuning is generally more performative \cite{lin2023videollavalearningunitedvisual,moon2023anymal}. Given these insights, we select AnyMAL~\cite{moon2023anymal} as a representative VCLM method for our benchmark tasks. AnyMAL uses InternVideo \cite{wang2022internvideo} as a video encoder, followed by an attention-based projector (perceiver resampler \cite{flamingo}) and is trained with both alignment training and instruction tuning. Furthermore, AnyMAL is trained on the HowTo100M dataset\cite{miech2019howto100m}. This dataset features videos of people performing daily tasks and is thus more relevant to our use case than VCLMs trained with other video datasets. Finally, AnyMAL also features 13-billion and 70-billion parameter versions, allowing us to test scaling laws for multimodal LLM prediction. The InternVideo endocder samples 8 frames from input videos. These frames, once encoded, use 256 tokens of the 2048 token context window of the Llama 2 models.
\begin{table*}[]
\centering
\scriptsize
\resizebox{\textwidth}{!}{
\begin{tabular}{lccccc}
\hline
\textbf{Method} & \textbf{Text}                      & \textbf{VLMs}                     & \textbf{Goal-Conditioned} & \textbf{Prediction} & \textbf{LLM}   \\ \hline
Palm\cite{huang2023palm}            & Actions, Narrations                & EgoVLP, Blip2                     & No                        & Direct              & GPT-Neo-1.3B   \\
Ant-GPT\cite{zhao2023antgpt}         & Actions                            & CLIP                              & Yes                       & Chain-of-thought    & Llama2 7B/13B  \\
VideoChat-Text\cite{li2023videochat}  & Actions, Objects, Audio Transcript & InternVideo, InternImage, Whisper & No                        & Direct              & Vicuna 13B     \\
Socratic (Ours) & Narration                          & LaViLa-Narrator                   & Yes                       & Direct              & Llama2 13B/70B \\ \hline
\end{tabular}}
\caption{\textbf{A comparison of Socratic models for prediction.} Models are compared by type of text, VLMs used, whether they are goal-conditioned, how they perform prediction, and which language models they use.}
\label{tab:socratic_models}
\end{table*}
\begin{table*}[]
\centering
\scriptsize
\resizebox{\textwidth}{!}{
\begin{tabular}{lccccc}
\hline
\textbf{Method} & \textbf{Encoder} & \textbf{Aggregation}  & \textbf{Alignment Training} & \textbf{Instruction Tuning} & \textbf{LLM}        \\ \hline
Video-LLaVA\cite{lin2023videollavalearningunitedvisual}     & LangugeBind      & N/A                   & Yes                         & Yes                         & Vicuna 7B           \\
LLaVA-NeXT\cite{liu2024llava}      & CLIP             & Concatenation         & Yes                         & Yes                         & Vicuna 7B/13B       \\
Video-ChatGPT\cite{maaz2023video}   & CLIP             & AveragePooling        & No                          & Yes                         & Vicuna 7B           \\
VideoChat-Embed\cite{li2023videochat} & BLIP2            & Q-Attention           & Yes                         & Yes                         & Vicuna 13B          \\
Video-Llama\cite{zhang2023video}     & BLIP2            & Q-Attention           & Yes                         & No                          & Llama2 7B/13B       \\
Video-Llama 2 \cite{cheng2024videollama2advancingspatialtemporal}   & CLIP             & Spatial-Temporal Conv & Yes                         & Yes                         & Mistral-Instruct 7B \\
TimeChat \cite{ren2024timechattimesensitivemultimodallarge}       & EVA-CLIP         & Q-Attention           & No                          & Yes                         & Llama2 7B           \\
AnyMAL (Ours) \cite{moon2023anymal}         & InternVideo      & N/A                   & Yes                         & Yes                         & Llama2 13B/70B      \\ \hline
\end{tabular}}
\caption{\textbf{A comparison of Visually-Conditioned Language Models capable of processing video.} Models are compared by vision encoder, how image level features are aggregated (if needed), whether the model is explicitly trained to align video features to text, and whether the model is instruction tuned.}
\label{tab:vclm_models}
\end{table*}

\section{Activity Scripts for Online Evaluation}
\label{sup:task_scripts}
The following are the scripts participants used in the online study. Users complete steps up to the ``\textbf{Get Assistance}" mark in any order they deem reasonable. The vision-based assistant takes over from there to guide users in completing the activities. The remaining steps are the steps we expect users to execute to successfully complete the activity.
\subsection{Prepare a Latte}
\begin{enumerate}
    \item Get a cup and put it in the espresso machine
    \item Pull 2x espresso shot using the espresso machine
    \item Pour milk into a metal pitcher\\
        \textbf{Get Assistance}
    \item Froth milk using the steam wand
    \item Pour milk into espresso cup
\end{enumerate}
\subsection{Make a Caprese Salad}
\begin{enumerate}
    \item Cut the tomato into slices
    \item Cut the fresh mozzarella into slices
    \item Tear the basil leaves
    \item Arrange the tomato on the plate\\
        \textbf{Get Assistance}
    \item Arrange the mozzarella slices on the plate
    \item Sprinkle the torn basil on top 
    \item Drizzle olive oil on top
\end{enumerate}
\subsection{Make a BLT Sandwich}
\begin{enumerate}
    \item Cut three slices of tomato
    \item Pull off a leaf of lettuce
    \item Take two slices of bread
    \item Put mayonnaise on the bottom piece of bread
    \item Put lettuce on the bottom piece of bread\\
        \textbf{Get Assistance}
    \item Put tomato slices on top of the lettuce
    \item Put bacon on top of the tomato slices
    \item Put the top piece of bread on

\end{enumerate}
\section{Model Error Analysis in Online Evaluation}
\label{sup:error_analysis}

\begin{table}[]
\resizebox{\columnwidth}{!}{
\begin{tabular}{lcccc}
\toprule
\textbf{Method}   & \multicolumn{1}{c}{\textbf{Task}} & \textbf{Redundant} & \textbf{Infeasible} & \textbf{Irrelevant} \\ 
\toprule
\multirow{4}{*}{{VCLM}}     & BLT                       & 7         & 4          & 2          \\
         & Caprese                   & 17        & 3          & 1          \\
         & Latte                     & 8         & 6          & 1          \\ 
         \cmidrule(l){2-5} 
         & Total                     & 32        & 13         & 4          \\ 
         \midrule
\multirow{4}{*}{{Socratic}} & BLT                       & 16        & 4          & 3          \\
         & Caprese                   & 12        & 0          & 1          \\
         & Latte                     & 5         & 12         & 3          \\ 
         \cmidrule(l){2-5} 
         & Total                     & 33        & 16         & 7          \\ 
\bottomrule
\end{tabular}}
\vspace{.2cm}
\caption{\textbf{A breakdown of cases where participants skipped assistant instructions}. Skips were categorized as redundant actions the participant had already completed, infeasible actions that could not be completed in the current task, and irrelevant actions which had no bearing on the current task. The numbers represent total numer of skips across all participants.}
\label{tab:mistakes}
\end{table}
 Table~\ref{tab:mistakes} shows a detailed breakdown of cases when participants skipped assistant instructions by method and activity. Recall that participants could skip instructions that were already completed (redundant), were infeasible in the current activity setting, or were irrelevant to the activity at hand. Both the Socratic and VCLM approaches have a similar total amount of skipped instructions and a similar distribution across skip reasons. While redundant skips were the most common type of skip by far, (63\% of all skips) the Socratic approach suggested a lot of infeasible actions for the latte activity specifically. Many of these actions were relevant to other latte settings (grinding coffee beans) or actions that would have been completed prior to the start of the activity episode (like setting up the espresso machine). Interestingly, the VCLM approach made significantly fewer infeasible suggestions for the latte task, which may be a product of its direct visual conditioning. However, this lower skip rate did not translate to a higher activity completion rate.
\section{Study Activity Visualization}
\begin{figure*}[t]
    \centering
    \includegraphics[width=1\textwidth]{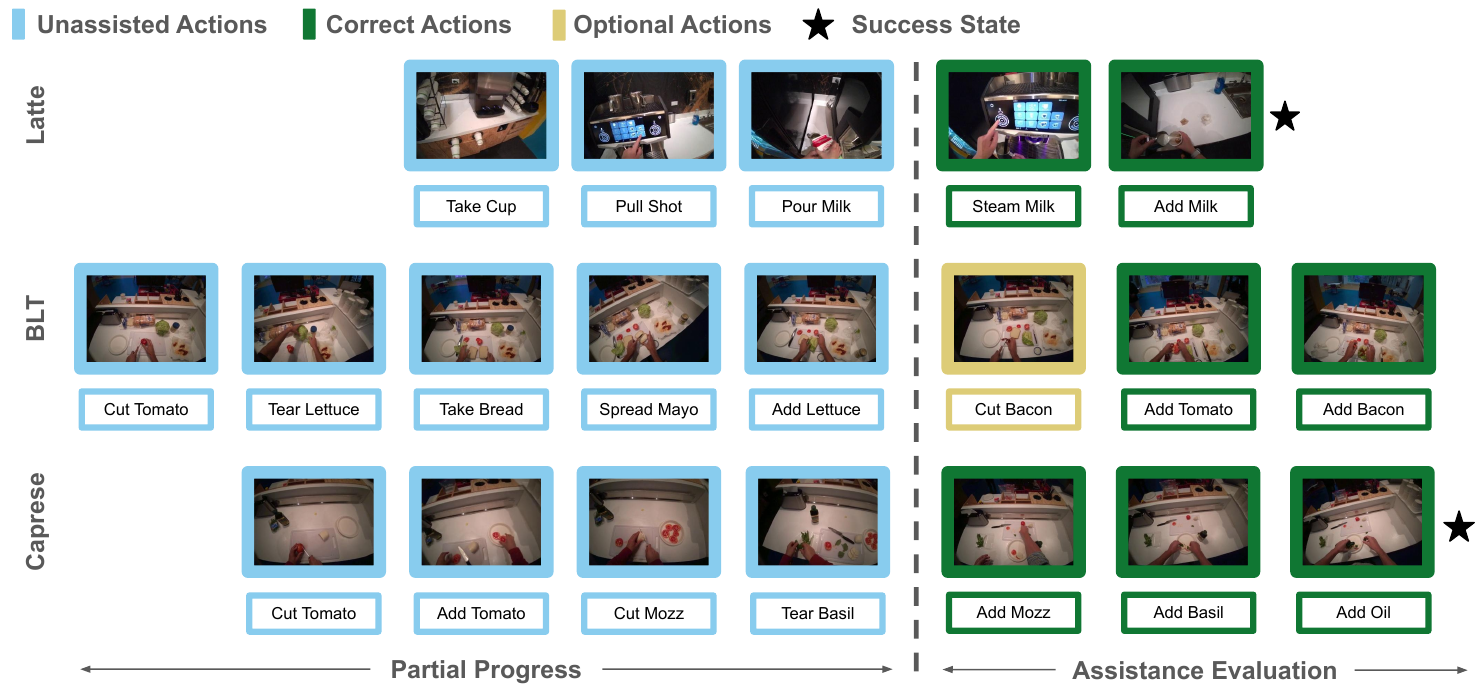}
    \caption{\textbf{A visualization of the most successful episodes from each cooking activity.} From top to bottom the activities are: make an espresso latte, make a BLT sandwich, make a caprese salad. No participant successfully accomplished the BLT activity in our study.}
    \vspace{-0.3cm}
    \label{fig:study_vis}
\end{figure*}

\begin{figure*}[t]
    \centering
    \includegraphics[width=1\textwidth]{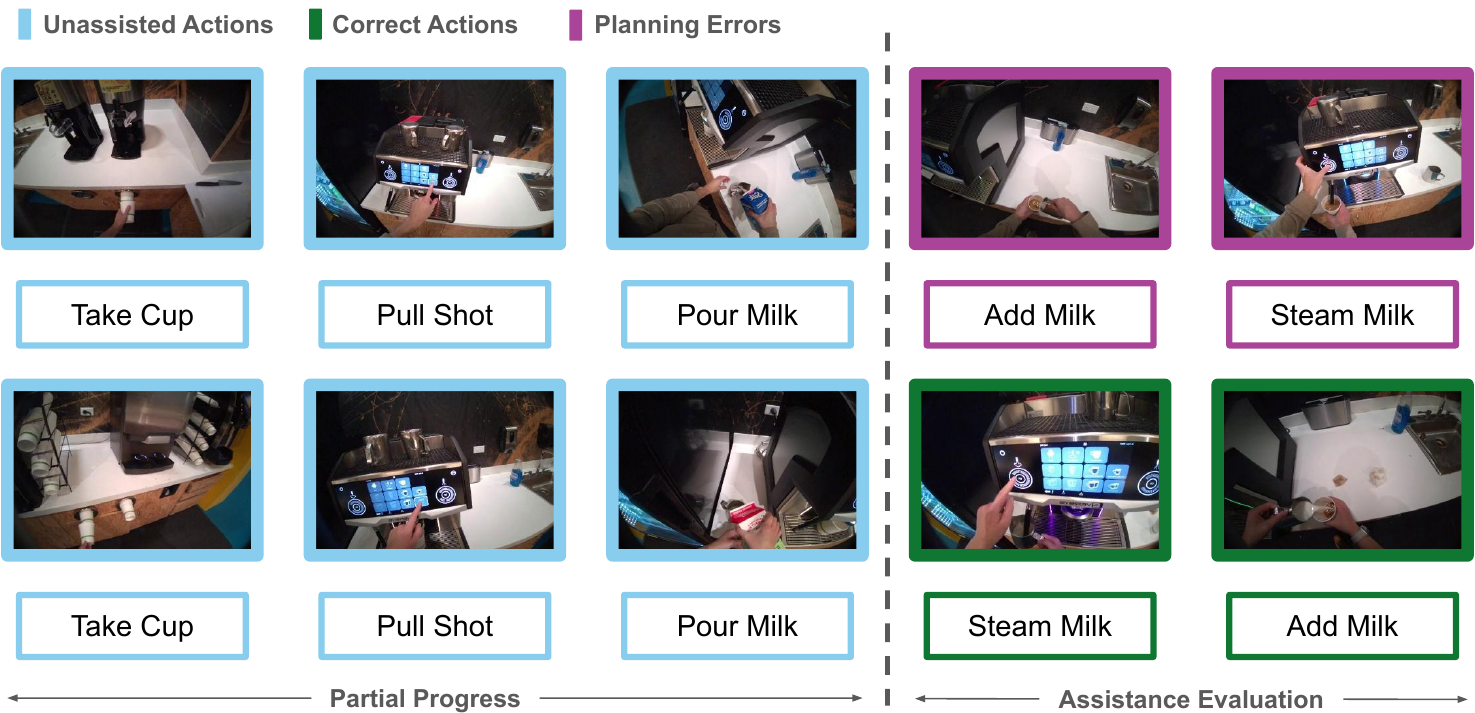}
    \caption{\textbf{Visualization of a planning mistake in the latte activity}. Top row shows a planning mistake during the latte activity. Bottom row shows the correct sequence of actions for completing the activity. Optional actions are omitted from the sequence.}
    \vspace{-0.3cm}
    \label{fig:latte_mistake_vis}
\end{figure*}

Figure~\ref{fig:study_vis} visualizes the most successful episode from each of the 3 cooking activities in the online study. Unfortunately, the BLT sandwich activity had no successful episodes so the closest episode is visualized. Figure~\ref{fig:latte_mistake_vis} shows a planning mistake in the latte activity. In the top row, the assistant suggested that the participant add milk before steaming it. The bottom row shows the correct sequence of actions. Grounding mistakes result in skipped actions and were not executed.

\section{Participant Data Collection Practices}
We obtained internal approval to collect egocentric videos from volunteer study participants in our office. We ensured that egocentric data contained no identifiable participant information. The data was stored in a private drive which only the study administrators had access to. The faces of other people who appear in participant videos were also blurred to protect their identities. We have no plans to release the full data beyond what is currently available for visualization purposes.

\end{document}